\documentclass[fleqn,10pt]{wlscirep}
\usepackage[utf8]{inputenc}
\usepackage[T1]{fontenc}
\usepackage{lineno}
\usepackage[symbol]{footmisc}
\usepackage{multirow}

\newcommand{\ie}{\textit{i.e.}}
\newcommand{\eg}{\textit{e.g.}}

\title{MedMNIST v2 - A large-scale lightweight benchmark for 2D and 3D biomedical image classification}

\author[1]{Jiancheng Yang} 
\author[1]{Rui Shi} 
\author[2]{Donglai Wei} 
\author[3]{Zequan Liu} 
\author[4]{Lin Zhao} 
\author[5]{Bilian Ke} 
\author[6]{Hanspeter Pfister} 
\author[1]{Bingbing Ni} 
\affil[1]{Shanghai Jiao Tong University, Shanghai, China}
\affil[2]{Boston College, Chestnut Hill, MA}
\affil[3]{RWTH Aachen University, Aachen, Germany}
\affil[4]{Department of Endocrinology and Metabolism, Fudan Institute of Metabolic Diseases, Zhongshan Hospital, Fudan University, Shanghai, China}
\affil[5]{Department of Ophthalmology, Shanghai General Hospital, Shanghai Jiao Tong University School of Medicine, Shanghai, China}
\affil[6]{Harvard University, Cambridge, MA}

\affil[*]{corresponding author(s): Bingbing Ni (nibingbing@sjtu.edu.cn)}

\begin{abstract}
We introduce \emph{MedMNIST v2}, a large-scale MNIST-like dataset collection of standardized biomedical images, including 12 datasets for 2D and 6 datasets for 3D. All images are pre-processed into a small size of $28\times 28$ (2D) or $28\times 28\times 28$ (3D) with the corresponding classification labels so that no background knowledge is required for users. Covering primary data modalities in biomedical images, MedMNIST v2 is designed to perform classification on lightweight 2D and 3D images with various dataset scales (from 100 to 100,000) and diverse tasks (binary/multi-class, ordinal regression, and multi-label). The resulting dataset, consisting of 708,069 2D images and 9,998 3D images in total, could support numerous research / educational purposes in biomedical image analysis, computer vision, and machine learning. We benchmark several baseline methods on MedMNIST v2, including 2D / 3D neural networks and open-source / commercial AutoML tools. The data and code are publicly available at \url{https://medmnist.com/}.

\end{abstract}
\begin{document}

\flushbottom
\maketitle

\section*{Background \& Summary}

\begin{figure}[!htb]
\centering
\includegraphics[width=\linewidth]{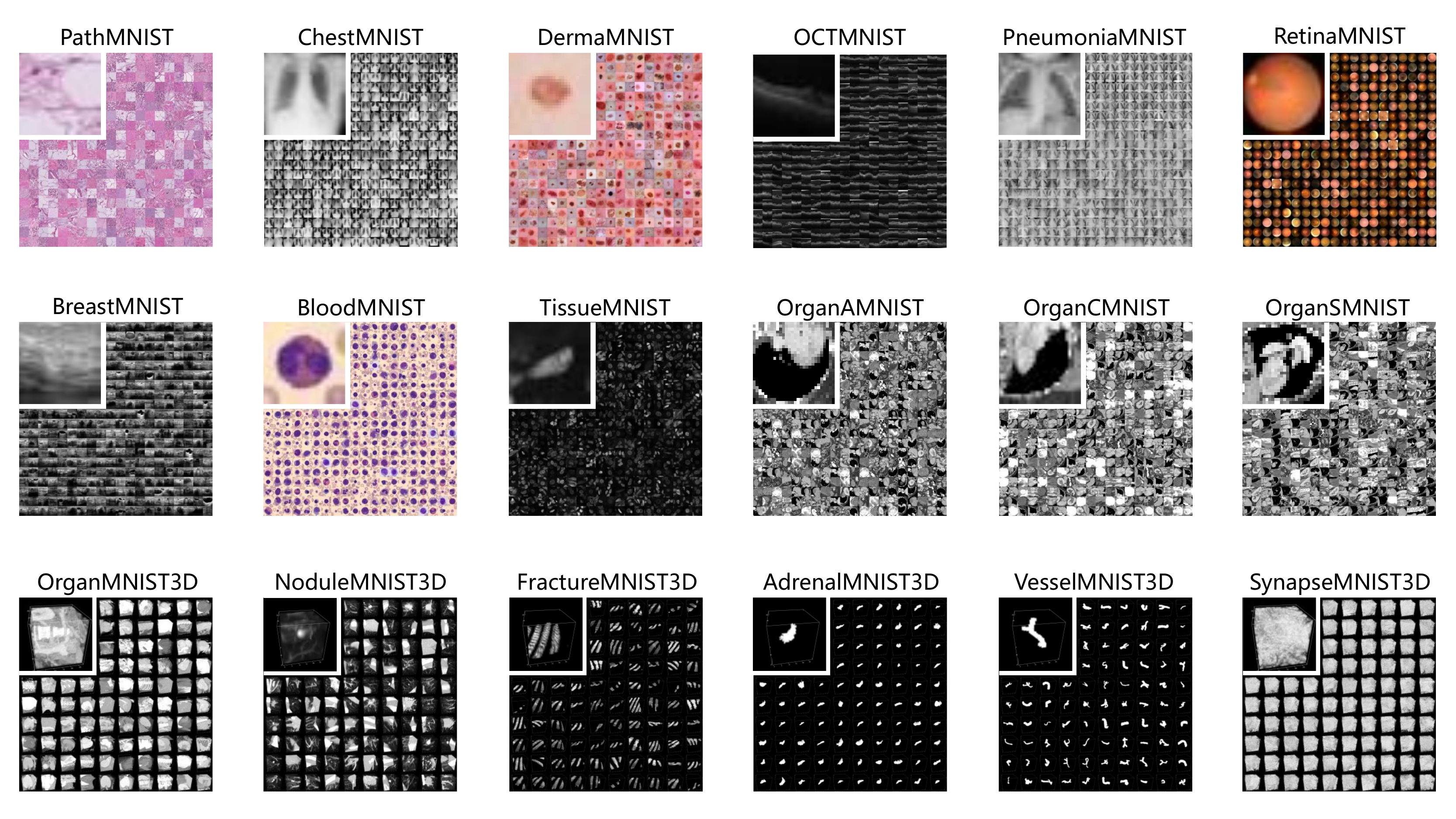}
\caption{\textbf{An overview of MedMNIST v2.} MedMNIST is a large-scale MNIST-like collection of standardized 2D and 3D biomedical images with classification labels. It is designed to be diverse, standardized, educational, and lightweight, which could support numerous research / educational purposes.}
\label{fig:medmnist_overview}
\end{figure}


Deep learning based biomedical image analysis plays an important role in the intersection of artificial intelligence and healthcare~\cite{shen2017deep,litjens2017survey,liu2019comparison}. Is deep learning a panacea in this area? Because of the inherent complexity in biomedicine, data modalities, dataset scales and tasks in biomedical image analysis could be highly diverse. Numerous biomedical imaging modalities are designed for specific purposes by adjusting sensors and imaging protocols. The biomedical image dataset scales in biomedical image analysis could range from 100 to 100,000. Moreover, even only considering medical image classification, there are binary/multi-class classification, multi-label classification, and ordinal regression. As a result, it needs large amounts of engineering effort to tune the deep learning models in real practice. On the other hand, it is not easy to identify whether a specific model design could be generalizable if it is only evaluated on a few datasets. Large and diverse datasets are urged by the research communities to fairly evaluate generalization performance of models.

Benchmarking data-driven approaches on various domains has been addressed by researchers. Visual Domain Decathlon (VDD)~\cite{rebuffi2017learning} develops an evaluation protocol on 10 existing natural image datasets to assess the model generalizability on different domains. In medical imaging area, Medical Segmentation Decathlon (MSD)~\cite{simpson2019large} introduces 10 3D medical image segmentation datasets to evaluate end-to-end segmentation performance: from whole 3D volumes to targets. It is
particularly important to understand the end-to-end performance of the current state of the art with MSD. However, the contribution of each part in the end-to-end systems could be particularly hard to analyze. As reported in the winning solutions~\cite{antonelli2021medical,isensee2021nnu}, hyperparameter tuning, pre/post-processing, model ensemble strategies and training/test-time augmentation could be more important than the machine learning part (\eg, model architectures, learning scheme). Therefore, a large but simple dataset focusing on the machine learning part like VDD, rather than the end-to-end system like MSD, will serve as a better benchmark to evaluate the generalization performance of the machine learning algorithms on the medical image analysis tasks.

In this study, we aim at a new ``decathlon'' dataset for biomedical image analysis, named \emph{MedMNIST v2}. As illustrated in Figure~\ref{fig:medmnist_overview}, MedMNIST v2 is a large-scale benchmark for 2D and 3D biomedical image classification, covering 12 2D datasets with 708,069 images and 6 3D datasets with 9,998 images. It is designed to be:

\begin{itemize}

    \item \textbf{Diverse}: It covers diverse data modalities, dataset scales (from 100 to 100,000), and tasks (binary/multi-class, multi-label, and ordinal regression). It is as diverse as the VDD~\cite{rebuffi2017learning} and MSD~\cite{simpson2019large} to fairly evaluate the generalizable performance of machine learning algorithms in different settings, but both 2D and 3D biomedical images are provided. 
    
    \item \textbf{Standardized}: Each sub-dataset is pre-processed into the same format (see details in Methods), which requires no background knowledge for users. As an MNIST-like~\cite{lecun2010mnist} dataset collection to perform classification tasks on small images, it primarily focuses on the machine learning part rather than the end-to-end system. Furthermore, we provide standard train-validation-test splits for all datasets in MedMNIST v2, therefore algorithms could be easily compared.
    
    \item \textbf{Lightweight}: The small size of $28\times 28$ (2D) or $28\times 28\times 28$ (3D) is friendly to evaluate machine learning algorithms.
    
    \item \textbf{Educational}: As an interdisciplinary research area, biomedical image analysis is difficult to hand on for researchers from other communities, as it requires background knowledge from computer vision, machine learning, biomedical imaging, and clinical science. Our data with the Creative Commons (CC) License is easy to use for educational purposes.
\end{itemize}

MedMNIST v2 is extended from our preliminary version, MedMNIST v1~\cite{medmnistv1}, with 10 2D datasets for medical image classification. As MedMNIST v1 is more medical-oriented, we additionally provide 2 2D bioimage datasets. Considering the popularity of 3D imaging in biomedical area, we carefully develop 6 3D datasets following the same design principle as 2D ones. A comparison of the ``decathlon'' datasets could be found in Table~\ref{tab:decathlon-comparison}.  We benchmark several standard deep learning methods and AutoML tools with MedMNIST v2 on both 2D and 3D datasets, including ResNets~\cite{he2016deep} with early-stopping strategies on validation set, open-source AutoML tools (auto-sklearn~\cite{NIPS2015_5872} and AutoKeras~\cite{jin2019auto}) and a commercial AutoML tool, Google AutoML Vision (for 2D only). All benchmark experiments are repeated at least 3 times for more stable results than in MedMNIST v1. Besides, the code for MedMNIST has been refactored to make it more friendly to use.

As a large-scale benchmark in biomedical image analysis, MedMNIST has been particularly useful for machine learning and computer vision research~\cite{qi2021elastic,chen2021alleviating,henn2021principled}, \eg, AutoML, trustworthy machine learning, domain adaptive learning. Moreover, considering the scarcity of 3D image classification datasets, the MedMNIST3D in MedMNIST v2 from diverse backgrounds could benefit research in 3D computer vision.

\begin{table*}[!htb]
\scriptsize
\centering
\caption{{\bf A comparison of MedMNIST v2 and other ``decathlon'' datasets}.}
\label{tab:decathlon-comparison}
\begin{tabular}{lcccc}
\toprule
& Visual Domain Decathlon~\cite{rebuffi2017learning} & Medical Segmentation Decathlon~\cite{simpson2019large} & MedMNIST v1~\cite{medmnistv1} & MedMNIST v2 \\ 
\midrule
Domain & Natural & Medical & Medical & Medical\\ 
Task & Classification & Segmentation & Classification & Classification  \\
Datasets & 10 & 10 & 10 & 18\\
2D / 3D & 2D & 3D & 2D & 2D \& 3D\\
Image Size & Variable ($\approx 72^2$) & Variable ($\approx(30-300)^3$) & Fixed ($28^2$)& Fixed ($28^2$ \& $28^3$)\\

\bottomrule
\end{tabular}
\end{table*}

\section*{Methods} \label{sec:methods}

\subsection*{Design Principles}

\begin{figure}[!htb]
\centering
\includegraphics[width=.95\linewidth]{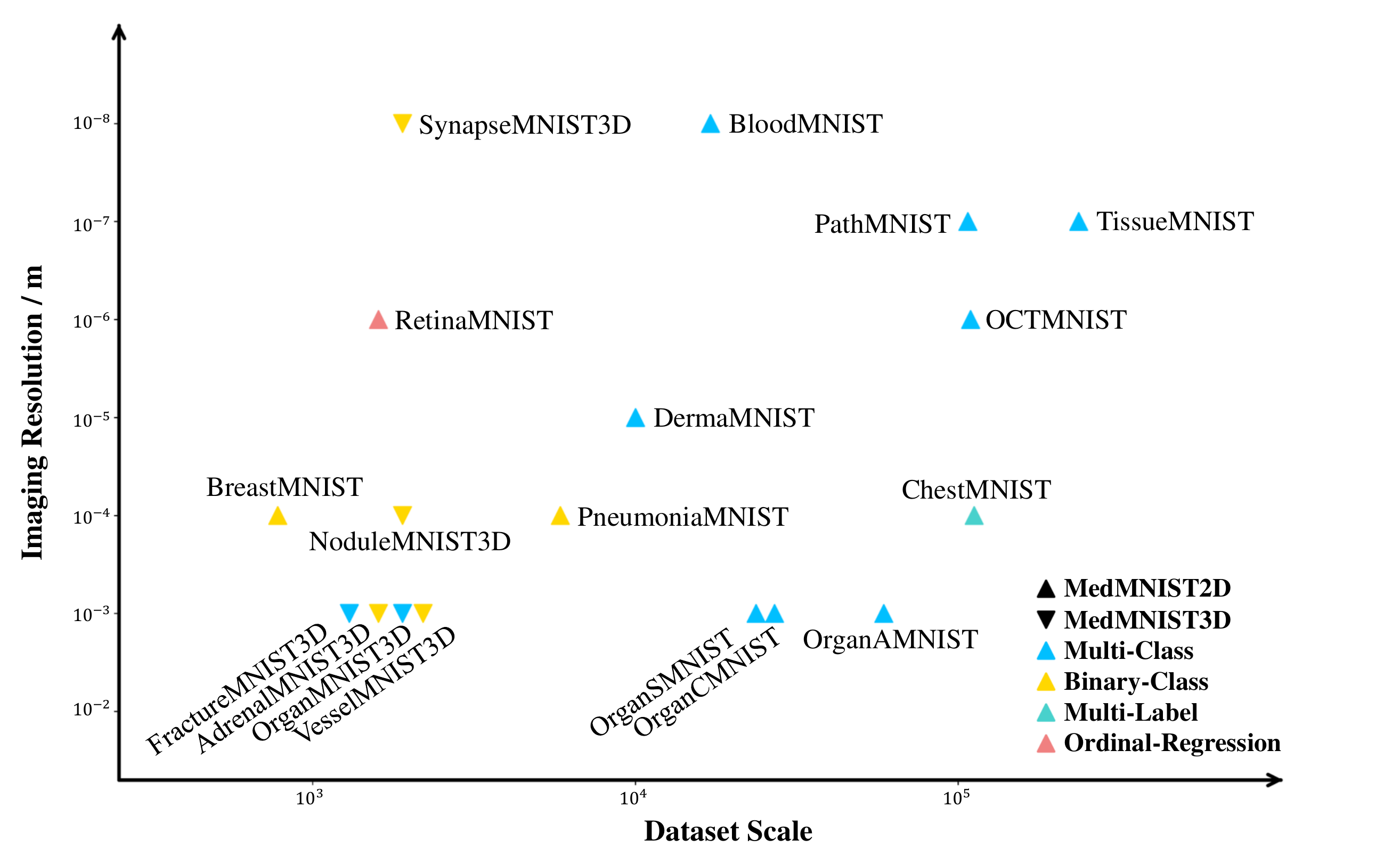}
\caption{\textbf{The landscape of MedMNIST v2.} The horizontal axis denotes the base-10 logarithm of the dataset scale, and the vertical axis denotes base-10 logarithm of imaging resolution. The upward and downward triangles are used to distinguish between 2D datasets and 3D datasets, and the 4 different colors represent different tasks.}
\label{fig:medmnist-landscape}
\end{figure}

The MedMNIST v2 dataset consists of 12 2D and 6 3D standardized datasets from carefully selected sources covering primary data modalities (\eg, X-ray, OCT, ultrasound, CT, electron microscope), diverse classification tasks (binary/multi-class, ordinal regression, and multi-label) and dataset scales (from 100 to 100,000). We illustrate the landscape of MedMNIST v2 in Figure~\ref{fig:medmnist-landscape}. As it is hard to categorize the data modalities, we use the imaging resolution instead to represent the modality. The diverse dataset design could lead to diverse task difficulty, which is desirable as a biomedical image classification benchmark. 

Although it is fair to compare performance on the test set only, it could be expensive to compare the impact of the train-validation split. Therefore, we provide an official train-validation-test split for each subset. We use the official data split from source dataset (if provided) to avoid data leakage. If the source dataset has only a split of training and validation set, we use the official validation set as test set and split the official training set with a ratio of 9:1 into training-validation. For the dataset without an official split, we split the dataset randomly at the patient level with a ratio of 7:1:2 into training-validation-test. All images are pre-processed into a MNIST-like format, \ie,  $28\times 28$ (2D) or $28\times 28\times 28$ (3D), with cubic spline interpolation operation for image resizing. The MedMNIST uses the classification labels from the source datasets directly in most cases, but the labels could be simplified (merged or deleted classes) if the classification tasks on the small images are too difficult. All source datasets are either associated with the Creative Commons (CC) Licenses or developed by us, which allows us to develop derivative datasets based on them. Some datasets are under CC-BY-NC license; we have contacted the authors and obtained the permission to re-distribute the datasets.

We list the details of all datasets in Table~\ref{tab:overview}. For simplicity, we call the collection of all 2D datasets as MedMNIST2D, and that of 3D as MedMNIST3D. In the next sections, we will describe how each dataset is created. 

\begin{table*}[!htb]
\scriptsize
\centering
\caption{{\bf Data summary of MedMNIST v2 dataset}, including data source, data modality, type of the classification task together with the number of classes for multi-class or that of labels for multi-label, number of samples in total and in each data split (training/validation/test). Upper: MedMNIST2D, 12 datasets of 2D images. Lower: MedMNIST3D, 6 datasets of 3D images. MC: Multi-Class. BC: Binary-Class. ML: Multi-Label. OR: Ordinal Regression.}
\label{tab:overview}
\begin{tabular}{lccccc}
\toprule
Name & Source & Data Modality & Task (\# Classes / Labels) & \# Samples & \# Training / Validation / Test \\ 
\midrule
\textit{MedMNIST2D}\\
PathMNIST & Kather et al.~\cite{10.1371/journal.pmed.1002730, kather_jakob_nikolas_2018_1214456} & Colon Pathology & MC (9) & 107,180 & 89,996 / 10,004 / 7,180\\
ChestMNIST & Wang et al.~\cite{wang2017chestxray} & Chest X-Ray & ML (14) BC (2) & 112,120 & 78,468 / 11,219 / 22,433  \\
DermaMNIST & Tschandl et al.~\cite{DBLP:journals/corr/abs-1803-10417, DVN/DBW86T_2018}, Codella et al.~\cite{codella2019skin} & Dermatoscope & MC (7) & 10,015 & 7,007 / 1,003 / 2,005 \\
OCTMNIST & Kermany et al.~\cite{KERMANY20181122, octmnist_dataset} & Retinal OCT & MC (4) & 109,309 & 97,477 / 10,832 / 1,000 \\
PneumoniaMNIST & Kermany et al.~\cite{KERMANY20181122, octmnist_dataset} & Chest X-Ray & BC (2) & 5,856 &  4,708 / 524 / 624 \\
RetinaMNIST & DeepDRiD Team\cite{deepdr} & Fundus Camera & OR (5) & 1,600 & 1,080 / 120 / 400  \\
BreastMNIST & Al-Dhabyani et al.~\cite{ALDHABYANI2020104863} & Breast Ultrasound & BC (2) & 780 & 546 / 78 / 156  \\
BloodMNIST & Acevedo et al.~\cite{ACEVEDO2020105474, bloodmnist_dataset} & Blood Cell Microscope & MC (8) & 17,092 & 11,959 / 1,712 / 3,421 \\
TissueMNIST & Ljosa et al.~\cite{ljosa2012annotated} & Kidney Cortex Microscope & MC (8) & 236,386 & 165,466 / 23,640 / 47,280  \\
OrganAMNIST & Bilic et al.~\cite{DBLP:journals/corr/abs-1901-04056}, Xu et al.~\cite{8625393} & Abdominal CT & MC (11) & 58,850 & 34,581 / 6,491 / 17,778 \\
OrganCMNIST & Bilic et al.~\cite{DBLP:journals/corr/abs-1901-04056}, Xu et al.~\cite{8625393} & Abdominal CT & MC (11) & 23,660 & 13,000 / 2,392 / 8,268 \\
OrganSMNIST & Bilic et al.~\cite{DBLP:journals/corr/abs-1901-04056}, Xu et al.~\cite{8625393} & Abdominal CT & MC (11) & 25,221 & 13,940 / 2,452 / 8,829  \\
\midrule
\textit{MedMNIST3D}\\
OrganMNIST3D & Bilic et al.~\cite{DBLP:journals/corr/abs-1901-04056}, Xu et al.~\cite{8625393} & Abdominal CT & MC (11) & 1,743 & 972 / 161 / 610 \\
NoduleMNIST3D & Armato et al.~\cite{https://doi.org/10.1118/1.3528204} & Chest CT & BC (2) & 1,633 & 1,158 / 165 / 310 \\
AdrenalMNIST3D & New & Shape from Abdominal CT & BC (2) & 1,584 & 1,188 / 98 / 298\\
FractureMNIST3D & Jin et al.~\cite{JIN2020103106}& Chest CT & MC (3) & 1,370 & 1,027 / 103 / 240 \\
VesselMNIST3D & Yang et al.~\cite{Yang_2020_CVPR} & Shape from Brain MRA & BC (2) & 1,909 & 1,335 / 192 / 382 \\
SynapseMNIST3D & New & Electron Microscope & BC (2) & 1,759 & 1,230 / 177 / 352 \\ 
\bottomrule
\end{tabular}
\end{table*}

\subsection*{Details for MedMNIST2D}

\subsubsection*{PathMNIST}

The PathMNIST is based on a prior study~\cite{10.1371/journal.pmed.1002730, kather_jakob_nikolas_2018_1214456} for predicting survival from colorectal cancer histology slides, providing a dataset (NCT-CRC-HE-100K) of $ 100,000 $ non-overlapping image patches from hematoxylin \& eosin stained histological images, and a test dataset (CRC-VAL-HE-7K) of $ 7,180 $ image patches from a different clinical center. The dataset is comprised of 9 types of tissues, resulting in a multi-class classification task. We resize the source images of $3 \times 224 \times 224$ into $3 \times 28 \times 28$, and split NCT-CRC-HE-100K into training and validation set with a ratio of $9:1$. The CRC-VAL-HE-7K is treated as the test set. 

\subsubsection*{ChestMNIST}
The ChestMNIST is based on the NIH-ChestXray14 dataset~\cite{wang2017chestxray}, a dataset comprising $ 112,120 $ frontal-view X-Ray images of $ 30,805 $ unique patients with the text-mined 14 disease labels, which could be formulized as a multi-label binary-class classification task. We use the official data split, and resize the source images of $1 \times 1,024 \times 1,024$ into $1 \times 28 \times 28$.

\subsubsection*{DermaMNIST}
The DermaMNIST is based on the HAM10000~\cite{DBLP:journals/corr/abs-1803-10417,codella2019skin,DVN/DBW86T_2018}, a large collection of multi-source dermatoscopic images of common pigmented skin lesions. The dataset consists of $ 10,015 $ dermatoscopic images categorized as 7 different diseases, formulized as a multi-class classification task. 
We split the images into training, validation and test set with a ratio of $ 7:1:2 $. The source images of $ 3 \times 600 \times 450 $ are resized into $ 3 \times 28 \times 28 $.

\subsubsection*{OCTMNIST}
The OCTMNIST is based on a prior dataset~\cite{KERMANY20181122, octmnist_dataset} of $109,309$ valid optical coherence tomography (OCT) images for retinal diseases. The dataset is comprised of 4 diagnosis categories, leading to a multi-class classification task. We split the source training set with a ratio of $ 9:1 $ into training and validation set, and use its source validation set as the test set. The source images are gray-scale, and their sizes are $ (384-1,536) \times (277-512) $. We center-crop the images with a window size of length of the short edge and resize them into $ 1 \times 28 \times 28 $.

\subsubsection*{PneumoniaMNIST}
The PneumoniaMNIST is based on a prior dataset~\cite{KERMANY20181122, octmnist_dataset} of $ 5,856 $ pediatric chest X-Ray images. The task is binary-class classification of pneumonia against normal. We split the source training set with a ratio of $ 9:1 $ into training and validation set, and use its source validation set as the test set. The source images are gray-scale, and their sizes are $ (384-2,916) \times (127-2,713) $. We center-crop the images with a window size of length of the short edge and resize them into $ 1 \times 28 \times 28 $.

\subsubsection*{RetinaMNIST}
The RetinaMNIST is based on the DeepDRiD~\cite{deepdr} challenge, which provides a dataset of $1,600$ retina fundus images. The task is ordinal regression for 5-level grading of diabetic retinopathy severity. We split the source training set with a ratio of $ 9:1 $ into training and validation set, and use the source validation set as the test set. The source images of $ 3 \times 1,736 \times 1,824 $ are center-cropped with a window size of length of the short edge and resized into $ 3 \times 28 \times 28 $.

\subsubsection*{BreastMNIST}
The BreastMNIST is based on a dataset~\cite{ALDHABYANI2020104863} of 780 breast ultrasound images. It is categorized into 3 classes: normal, benign, and malignant. As we use low-resolution images, we simplify the task into binary classification by combining normal and benign as positive and classifying them against malignant as negative. We split the source dataset with a ratio of $ 7:1:2 $ into training, validation and test set. The source images of $ 1 \times 500 \times 500 $ are resized into $ 1 \times 28 \times 28 $.

\subsubsection*{BloodMNIST}
The BloodMNIST is based on a dataset~\cite{ACEVEDO2020105474, bloodmnist_dataset} of individual normal cells, captured from individuals without infection, hematologic or oncologic disease and free of any pharmacologic treatment at the moment of blood collection. It contains a total of 17,092 images and is organized into 8 classes. We split the source dataset with a ratio of $ 7:1:2 $ into training, validation and test set. The source images with resolution $3 \times 360\times 363$ pixels are center-cropped into $ 3 \times 200\times 200$, and then resized into $3 \times 28 \times 28$.

\subsubsection*{TissueMNIST}
We use the BBBC051~\cite{woloshuk2020situ}, available from the Broad Bioimage Benchmark Collection~\cite{ljosa2012annotated}. The dataset contains $236,386$ human kidney cortex cells, segmented from 3 reference tissue specimens and organized into 8 categories. We split the source dataset with a ratio of $ 7:1:2 $ into training, validation and test set. Each gray-scale image is $32 \times 32 \times 7$ pixels, where $7$ denotes $7$ slices. We obtain 2D maximum projections by taking the maximum pixel value along the axial-axis of each pixel, and resize them into $28 \times 28$ gray-scale images.

\subsubsection*{Organ\{A,C,S\}MNIST} 
The {Organ\{A,C,S\}MNIST} is based on 3D computed tomography (CT) images from Liver Tumor Segmentation Benchmark (LiTS)~\cite{DBLP:journals/corr/abs-1901-04056}. They are renamed from OrganMNIST\_\{Axial,Coronal,Sagittal\} (in MedMNIST v1~\cite{medmnistv1}) for simplicity. We use bounding-box annotations of 11 body organs from another study~\cite{8625393} to obtain the organ labels. Hounsfield-Unit (HU) of the 3D images are transformed into gray-scale with an abdominal window. We crop 2D images from the center slices of the 3D bounding boxes in axial / coronal / sagittal views (planes). The only differences of Organ\{A,C,S\}MNIST are the views. The images are resized into $ 1 \times 28 \times 28 $ to perform multi-class classification of 11 body organs. 115 and 16 CT scans from the source training set are used as training and validation set, respectively. The 70 CT scans from the source test set are treated as the test set.

\subsection*{Details for MedMNIST3D}

\subsubsection*{OrganMNIST3D}
The source of the OrganMNIST3D is the same as that of the {Organ\{A,C,S\}MNIST}. Instead of 2D images, we directly use the 3D bounding boxes and process the images into $ 28 \times 28 \times 28 $ to perform multi-class classification of 11 body organs. The same 115 and 16 CT scans as the {Organ\{A,C,S\}MNIST} from the source training set are used as training and validation set, respectively, and the same 70 CT scans as the {Organ\{A,C,S\}MNIST} from the source test set are treated as the test set.

\subsubsection*{NoduleMNIST3D}
The NoduleMNIST3D is based on the LIDC-IDRI~\cite{https://doi.org/10.1118/1.3528204}, a large public lung nodule dataset, containing images from thoracic CT scans. The dataset is designed for both lung nodule segmentation and 5-level malignancy classification task. To perform binary classification, we categorize cases with malignancy level $1/2$ into negative class and $4/5$ into positive class, ignoring the cases with malignancy level $3$. We split the source dataset with a ratio of $ 7:1:2 $ into training, validation and test set, and center-crop the spatially normalized images (with a spacing of $1mm\times1mm\times1mm$) into $28 \times 28 \times 28$.

\subsubsection*{AdrenalMNIST3D}
The AdrenalMNIST3D is a new 3D shape classification dataset, consisting of shape masks from 1,584 left and right adrenal glands (\ie, 792 patients). Collected from Zhongshan Hospital Affiliated to Fudan University, each 3D shape of adrenal gland is annotated by an expert endocrinologist using abdominal computed tomography (CT), together with a binary classification label of normal adrenal gland or adrenal mass. Considering patient privacy, we do not provide the source CT scans, but the real 3D shapes of adrenal glands and their classification labels. We calculate the center of adrenal and resize the center-cropped $64mm \times 64mm \times 64mm$ volume into $28 \times 28 \times 28$. The dataset is randomly split into training / validation / test set of 1,188 / 98 / 298 on a patient level. 

\subsubsection*{FractureMNIST3D}
The FractureMNIST3D is based on the RibFrac Dataset~\cite{JIN2020103106}, containing around $5,000$ rib fractures from 660 computed tomography (CT) scans. The dataset organizes detected rib fractures into 4 clinical categories (\ie, buckle, nondisplaced, displaced, and segmental rib fractures). As we use low-resolution images, we disregard segmental rib fractures and classify 3 types of rib fractures (\ie, buckle, nondisplaced, and displaced). For each annotated fracture area, we calculate its center and resize the center-cropped $64mm \times 64mm \times 64mm$ image into $28 \times 28 \times 28$. The official split of training, validation and test set is used.

\subsubsection*{VesselMNIST3D}
The VesselMNIST3D is based on an open-access 3D intracranial aneurysm dataset, IntrA~\cite{Yang_2020_CVPR}, containing 103 3D models (meshes) of entire brain vessels collected by reconstructing MRA images. $1,694$ healthy vessel segments and $215$ aneurysm segments are generated automatically from the complete models. We fix the non-watertight mesh with PyMeshFix~\cite{Attene2010ALA} and voxelize the watertight mesh with trimesh~\cite{trimesh} into $28 \times 28 \times 28$ voxels. We split the source dataset with a ratio of $ 7:1:2 $ into training, validation and test set. 

\subsubsection*{SynapseMNIST3D}
The SynapseMNIST3D is a new 3D volume dataset to classify whether a synapse is excitatory or inhibitory. It uses a 3D image volume of an adult rat acquired by a multi-beam scanning electron microscope. The original data is of the size $100\times 100 \times 100 um^3$ and the resolution $8 \times 8 \times 30 nm^3$, where a $(30 um)^3$ sub-volume was used in the MitoEM dataset~\cite{wei2020mitoem} with dense 3D mitochondria instance segmentation labels. Three neuroscience experts segment a pyramidal neuron within the whole volume and proofread all the synapses on this neuron with excitatory / inhibitory labels. For each labeled synaptic location, we crop a 3D volume of $1024\times1024\times1024 nm^3$ and resize it into $28\times28\times28$ voxels. Finally, the dataset is randomly split with a ratio of $7:1:2$ into training, validation and test set.

\section*{Data Records}

The data files of MedMNIST v2 dataset can be accessed at Zenodo~\cite{medmnistv2zenodo}. It contains 12 pre-processed 2D datasets (MedMNIST2D) and 6 pre-processed 3D datasets (MedMNIST3D). Each subset is saved in NumPy~\cite{harris2020array} npz format, named as <data>mnist.npz for MedMNIST2D and <data>mnist3d.npz for MedMNIST3D, and is comprised of $6$ keys (``train\_images'', ``train\_labels'', ``val\_images'', ``val\_labels'', ``test\_images'', ``test\_labels''). The data type of the dataset is uint8.

\begin{itemize}
\item \textbf{``\{train,val,test\}\_images''}: an array containing images, with a shape of $N\times28\times28$ for 2D gray-scale datasets, of $N\times28\times28\times3$ for 2D RGB datasets, of $N\times28\times28\times28$ for 3D datasets. $N$ denotes the number of samples in training / validation / test set. 
\item \textbf{``\{train,val,test\}\_labels''}: an array containing ground-truth labels, with a shape of $N\times1$ for multi-class / binary-class / ordinal regression datasets, of $N\times L$ for multi-lable binary-class datasets. $N$ denotes the number of samples in training / validation / test set and $L$ denotes the number of task labels in the multi-label dataset (\ie, $14$ for the ChestMNIST). 
\end{itemize}

\section*{Technical Validation}

\subsection*{Baseline Methods}

For MedMNIST2D, we first implement ResNets~\cite{he2016deep} with a simple early-stopping strategy on validation set as baseline methods. The ResNet model contains 4 residual layers and each layer has several blocks, which is a stack of convolutional layers, batch normalization and ReLU activation. The input channel is always $3$ since we convert gray-scale images into RGB images. To fairly compare with other methods, the input resolutions are $28$ or $224$ (resized from $28$) for the ResNet-18 and ResNet-50. For all model training, we use cross entropy-loss and set the batch size as $128$. We utilize an Adam optimizer~\cite{kingma2014adam} with an initial learning rate of $0.001$ and train the model for $100$ epochs, delaying the learning rate by $0.1$ after $50$ and $75$ epochs. 

For MedMNIST3D, we implement ResNet-18 / ResNet-50\cite{he2016deep} with 2.5D / 3D / ACS~\cite{9314699} convolutions with a simple early-stopping strategy on validation set as baseline methods, using the one-line 2D neural network converters provided in the official ACS code repository (\url{https://github.com/M3DV/ACSConv}). When loading the datasets, we copy the single channel into 3 channels to make it compatible. For all model training, we use cross-entropy loss and set the batch size as $32$. We utilize an Adam optimizer~\cite{kingma2014adam} with an initial learning rate of $0.001$ and train the model for $100$ epochs, delaying the learning rate by $0.1$ after $50$ and $75$ epochs. Additionally, as a regularization for the two datasets of shape modality (\ie, AdrenalMNIST3D / VesselMNIST3D), we multiply the training set by a random value in $[0,1]$ during training and multiply the images by a fixed coefficient of $0.5$ during evaluation. 

The details of model implementation and training scheme can be found in our code.

\subsection*{AutoML Methods}

We have also selected several AutoML methods: auto-sklearn~\cite{NIPS2015_5872} as the representative of open-source AutoML tools for statistical machine learning, AutoKeras~\cite{jin2019auto} as the representative of open-source AutoML tools for deep learning, and Google AutoML Vision as the representative of commercial black-box AutoML tools, with deep learning empowered. We run auto-sklearn~\cite{NIPS2015_5872} and AutoKeras~\cite{jin2019auto} on both MedMNIST2D and MedMNIST3D, and Google AutoML Vision on MedMNIST2D only. 

\textbf{auto-sklearn}~\cite{NIPS2015_5872} automatically searches the algorithms and hyper-parameters in scikit-learn \cite{pedregosa2011scikit} package.
We set time limit for search of appropriate models according to the dataset scale. The time limit is 2 hours for 2D datasets with scale $<10,000$, 4 hours for those of $[10,000,50,000]$, and 6 hours for those $>50,000$. For 3D datasets, we set time limit as 4 hours. We flatten the images into one dimension, and provide reshaped one-dimensional data with the corresponding labels for auto-sklearn to fit.

\textbf{AutoKeras}~\cite{jin2019auto} based on Keras package~\cite{chollet2015keras} searches deep neural networks and hyper-parameters. For each dataset, we set number of max\_trials as 20 and number of epochs as 20. It tries 20 different Keras models and trains each model for 20 epochs. We choose the best model based on the highest AUC score on validation set.

\textbf{Google AutoML Vision} (\url{https://cloud.google.com/vision/automl/docs}, experimented in July, 2021) is a commercial AutoML tool offered as a service from Google Cloud. We train Edge exportable models of MedMNIST2D on Google AutoML Vision and export trained quantized models into TensorFlow Lite format to do offline inference. We set number of node hours of each dataset according to the data scale. We allocate $1$ node hour for dataset with scale around $1,000$, $2$ node hours for scale around $10,000$, $3$ node hours for scale around $100,000$, and $4$ node hours for scale around $200,000$. 

\subsection*{Evaluation}
Area under ROC curve (AUC)~\cite{bradley1997use} and Accuracy (ACC) are used as the evaluation metrics. AUC is a threshold-free metric to evaluate the continuous prediction scores, while ACC evaluates the discrete prediction labels given threshold (or $\arg\max$). AUC is less sensitive to class imbalance than ACC. Since there is no severe class imbalance on our datasets, ACC could also serve as a good metric. Although there are many other metrics, we simply select AUC and ACC for the sake of simplicity and standardization of evaluation. We report the AUC and ACC for each dataset. Data users are also encouraged to analyze the average performance over the $12$ 2D datasets and $6$ 3D datasets to benchmark their methods. Thereby, we report average AUC and ACC score over MedMNIST2D and MedMNIST3D respectively to easily compare the performance of different methods. 


\subsubsection*{Benchmark on Each Dataset}

The performance on each dataset of MedMNIST2D and MedMNIST3D is reported in Table~\ref{tab:2DResults} and Table~\ref{tab:3DResults}, respectively. We calculate the mean value of at least 3 trials for each method on each dataset. 

For 2D datasets, Google AutoML Vision is well-performing in general, however it could not always win, even compared with the baseline ResNet-18 and ResNet-50. Auto-sklearn performs poorly on most datasets, indicating that the typical statistical machine learning algorithms do not work well on our 2D medical image datasets. AutoKeras performs well on datasets with large scales, however relatively worse on datasets with small scale. With the same depth of ResNet backbone, datasets of resolution $224$ outperform resolution $28$ in general. For datasets of resolution $28$, ResNet-18 wins higher scores than ResNet-50 on most datasets. 

For 3D datasets, AutoKeras does not work well, while auto-sklearn performs better than on MedMNIST2D. Auto-sklearn is superior to ResNet-18+2.5D and ResNet-50+2.5D in general, and even outperforms all the other methods in ACC score on AdrenalMNIST3D. 2.5D models have poorer performance compared with 3D and ACS models, while 3D and ACS models are comparable to each other. With 3D convolution, ResNet-50 backbone surpasses ResNet-18.

\begin{table*}[!htb]
\footnotesize
	\caption{{\bf Benchmark on each dataset of MedMNIST2D} in metrics of AUC and ACC.}
	\label{tab:2DResults}
	\begin{center}
		
		\begin{tabular}{lcccccccccccc}
			\toprule
			\multirow{2}{*}{Methods} &
			\multicolumn{2}{c}{PathMNIST} &
			\multicolumn{2}{c}{ChestMNIST} &
			\multicolumn{2}{c}{DermaMNIST} &
			\multicolumn{2}{c}{OCTMNIST} &
			\multicolumn{2}{c}{PneumoniaMNIST} & 
			\multicolumn{2}{c}{RetinaMNIST} \\
			& AUC & ACC & AUC & ACC & AUC & ACC & AUC & ACC & AUC & ACC & AUC & ACC\\ \midrule
			ResNet-18 (28)~\cite{he2016deep}     & 0.983 & 0.907 & 0.768 & 0.947 & 0.917 & 0.735 & 0.943 & 0.743 & 0.944 & 0.854 & 0.717 & 0.524 \\
			ResNet-18 (224)~\cite{he2016deep}        & 0.989 & 0.909 & 0.773 & 0.947 & \bf 0.920 & 0.754 & 0.958 & 0.763 & 0.956 & 0.864 & 0.710 & 0.493 \\
			ResNet-50 (28)~\cite{he2016deep}         & \bf 0.990 & \bf 0.911 & 0.769 & 0.947 & 0.913 & 0.735 & 0.952 & 0.762 & 0.948 & 0.854 & 0.726 & 0.528  \\
			ResNet-50 (224)~\cite{he2016deep}        & 0.989 & 0.892 & 0.773 & \bf 0.948 & 0.912 & 0.731 & 0.958 & \bf 0.776 & 0.962 & 0.884 & 0.716 & 0.511 \\
			auto-sklearn~\cite{NIPS2015_5872}         & 0.934 & 0.716 & 0.649 & 0.779 & 0.902 & 0.719 & 0.887 & 0.601 & 0.942 & 0.855 & 0.690 & 0.515  \\
			AutoKeras~\cite{jin2019auto}           & 0.959 & 0.834 & 0.742 & 0.937 & 0.915 & 0.749 & 0.955 & 0.763 & 0.947 & 0.878 & 0.719 & 0.503 \\
			Google AutoML Vision  & 0.944 & 0.728 & \bf 0.778 & \bf 0.948 & 0.914 & \bf 0.768 & \bf 0.963 & 0.771 & \bf 0.991 & \bf 0.946 & \bf 0.750 & \bf 0.531 \\
			\midrule
			\multirow{2}{*}{Methods} &
			
			\multicolumn{2}{c}{BreastMNIST} &
			\multicolumn{2}{c}{BloodMNIST} &
			\multicolumn{2}{c}{TissueMNIST} &
			\multicolumn{2}{c}{OrganAMNIST} &
			\multicolumn{2}{c}{OrganCMNIST} &
			\multicolumn{2}{c}{OrganSMNIST} \\
			 & AUC & ACC & AUC & ACC & AUC & ACC & AUC & ACC & AUC & ACC & AUC & ACC \\ \midrule
			ResNet-18 (28)~\cite{he2016deep}          & 0.901 & \bf 0.863 & \bf 0.998 & 0.958 & 0.930 & 0.676 & 0.997 & 0.935 & 0.992 & 0.900 & 0.972 & 0.782 \\
			ResNet-18 (224)~\cite{he2016deep}         & 0.891 & 0.833 & \bf 0.998 & 0.963 & 0.933 & 0.681 & \bf 0.998 & \bf 0.951 & \bf 0.994 & \bf 0.920 & 0.974 & 0.778 \\
			ResNet-50 (28)~\cite{he2016deep}         & 0.857 & 0.812 & 0.997 & 0.956 & 0.931 & 0.680 & 0.997 & 0.935 & 0.992 & 0.905 & 0.972 & 0.770 \\
			ResNet-50 (224)~\cite{he2016deep}        & 0.866 & 0.842 & 0.997 & 0.950 & 0.932 & 0.680 & \bf 0.998 & 0.947 & 0.993 & 0.911 & \bf 0.975 & 0.785 \\
			auto-sklearn~\cite{NIPS2015_5872}        & 0.836 & 0.803 & 0.984 & 0.878 & 0.828 & 0.532 & 0.963 & 0.762 & 0.976 & 0.829 & 0.945 & 0.672 \\
			AutoKeras~\cite{jin2019auto}           & 0.871 & 0.831 & \bf 0.998 & 0.961 & \bf 0.941 & \bf 0.703 & 0.994 & 0.905 & 0.990 & 0.879 & 0.974 & \bf 0.813 \\
			Google AutoML Vision & \bf 0.919 & 0.861 & \bf 0.998 & \bf 0.966 & 0.924 & 0.673 & 0.990 & 0.886 & 0.988 & 0.877 & 0.964 & 0.749 \\
			\bottomrule
		\end{tabular}
	\end{center}
\end{table*}

\begin{table*}[!htb]
\scriptsize
	\caption{{\bf Benchmark on each dataset of MedMNIST3D} in metrics of AUC and ACC.}
	\label{tab:3DResults}
	\begin{center}
		
		\begin{tabular}{lcccccccccccc}
			\toprule
			\multirow{2}{*}{Methods} &
			\multicolumn{2}{c}{OrganMNIST3D} &
			\multicolumn{2}{c}{NoduleMNIST3D} &
			\multicolumn{2}{c}{FractureMNIST3D} &
			\multicolumn{2}{c}{AdrenalMNIST3D} &
			\multicolumn{2}{c}{VesselMNIST3D} &
			\multicolumn{2}{c}{SynapseMNIST3D}\\
			& AUC & ACC & AUC & ACC & AUC & ACC & AUC & ACC & AUC & ACC & AUC & ACC \\ \midrule
			ResNet-18~\cite{he2016deep}+2.5D & 0.977 & 0.788 & 0.838 & 0.835 & 0.587 & 0.451 & 0.718 & 0.772 & 0.748 & 0.846 & 0.634 & 0.696 \\
			
			ResNet-18~\cite{he2016deep}+3D & \bf 0.996 & \bf 0.907 & 0.863 & 0.844 & 0.712 & 0.508 & 0.827 & 0.721 & 0.874 & 0.877 & 0.820 & 0.745  \\
			ResNet-18~\cite{he2016deep}+ACS~\cite{9314699} & 0.994 & 0.900 & 0.873 & 0.847 & 0.714 & 0.497 & \bf 0.839 & 0.754 & \bf 0.930 & \bf 0.928 & 0.705 & 0.722  \\
			
			ResNet-50~\cite{he2016deep}+2.5D & 0.974 & 0.769 & 0.835 & 0.848 & 0.552 & 0.397 & 0.732 & 0.763 & 0.751 & 0.877 & 0.669 & 0.735 \\
			
			ResNet-50~\cite{he2016deep}+3D & 0.994 & 0.883 & 0.875 & 0.847 & 0.725 & 0.494 & 0.828 & 0.745 & 0.907 & 0.918 & \bf 0.851 & \bf 0.795 \\
			ResNet-50~\cite{he2016deep}+ACS~\cite{9314699} & 0.994 & 0.889 & 0.886 & 0.841 & \bf 0.750 & \bf 0.517 & 0.828 & 0.758 & 0.912 & 0.858 & 0.719 & 0.709 \\
			auto-sklearn~\cite{NIPS2015_5872} & 0.977 & 0.814 & \bf 0.914 & \bf 0.874 & 0.628 & 0.453 & 0.828 & \bf 0.802 & 0.910 & 0.915 & 0.631 & 0.730 \\
			AutoKeras~\cite{jin2019auto} & 0.979 & 0.804 & 0.844 & 0.834 & 0.642 & 0.458 & 0.804 & 0.705 & 0.773 & 0.894 & 0.538 & 0.724 \\
			\bottomrule
		\end{tabular}

	\end{center}
\end{table*}

\subsubsection*{Average Performance of Each Method}

To compare the performance of various methods, we report the average AUC and average ACC of each method over all datasets. The average performance of methods on MedMNIST2D and MedMNIST3D are reported in Table \ref{tab:AVG_2DResults} and Table \ref{tab:AVG_3DResults}, respectively. Despite the great gap among the metrics of different sub-datasets, the average AUC and ACC could still manifest the performance of each method. 

For MedMNIST2D, Google AutoML Vision outperforms all the other methods in average AUC, however, it is very close to the performance of baseline ResNets. The ResNets surpass auto-sklearn and AutoKeras, and outperform Google AutoML Vision in average ACC. Under the same backbone, the datasets with resolution of 224 win higher AUC and ACC score than resolution of 28. While under the same resolution, ResNet-18 is superior to ResNet-50. 

For MedMNIST3D, AutoKeras does not perform well, performing worse than auto-sklearn. Under the same ResNet backbone, 2.5D models are inferior to 3D and ACS models and perform worse than auto-sklearn and AutoKeras. Surprisingly, the ResNet-50 with standard 3D convolution outperforms all the other methods on average.

\begin{table*}[!htb]
\footnotesize
	\caption{{\bf Average performance of MedMNIST2D} in metrics of average AUC and average ACC over all 2D datasets.}
	\label{tab:AVG_2DResults}
	\begin{center}
		
		\begin{tabular}{lcc}
			\toprule
			Methods & AVG AUC & AVG ACC \\ 
			\midrule
			ResNet-18 (28)~\cite{he2016deep}     & 0.922 & 0.819  \\
			ResNet-18 (224)~\cite{he2016deep}        & 0.925 & \bf 0.821  \\
			ResNet-50 (28)~\cite{he2016deep}         & 0.920 & 0.816  \\
			ResNet-50 (224)~\cite{he2016deep}        & 0.923 & \bf 0.821  \\
			auto-sklearn~\cite{NIPS2015_5872}         & 0.878 & 0.722  \\
			AutoKeras~\cite{jin2019auto}           & 0.917 & 0.813  \\
			Google AutoML Vision  & \bf 0.927 & 0.809 \\
			\bottomrule
		\end{tabular}
	\end{center}
\end{table*}

\begin{table*}[!htb]
\footnotesize
	\caption{{\bf Average performance of MedMNIST3D} in metrics of average AUC and average ACC over all 3D datasets.}
	\label{tab:AVG_3DResults}
	\begin{center}
		
		\begin{tabular}{lcc}
			\toprule
			Methods & AVG AUC & AVG ACC \\ 
			\midrule
			ResNet-18~\cite{he2016deep}+2.5D   & 0.750 & 0.731  \\
			ResNet-18~\cite{he2016deep}+3D  & 0.849 & 0.767  \\  
			ResNet-18~\cite{he2016deep}+ACS~\cite{9314699}  & 0.842 & 0.775  \\
			ResNet-50~\cite{he2016deep}+2.5D   & 0.752 & 0.732  \\
			ResNet-50~\cite{he2016deep}+3D        & \bf 0.863 & \bf 0.780 \\
			ResNet-50~\cite{he2016deep}+ACS~\cite{9314699}  &  0.848 & 0.762  \\
			auto-sklearn~\cite{NIPS2015_5872}         & 0.815 & 0.765  \\
			AutoKeras~\cite{jin2019auto}           & 0.763 & 0.737  \\
			\bottomrule
		\end{tabular}
	\end{center}
\end{table*}

\subsubsection*{Difference between Organ\{A,C,S\}MNIST and OrganMNIST3D}
Organ\{A,C,S\}MNIST and OrganMNIST3D are generated from the same source dataset, and share the same task and the same data split. However, samples in the 2D and 3D datasets are different. Organ\{A,C,S\}MNIST are sampled slices of 3D bounding boxes of 3D CT images in axial / coronal / sagittal views (planes), respectively. They are sliced before being resized into $1 \times 28 \times 28$. On the other hand, OrganMNIST3D is resized into $28 \times 28 \times 28$ directly. Therefore, the Organ\{A,C,S\}MNIST metrics in Table~\ref{tab:2DResults} and the OrganMNIST3D metrics in Table~\ref{tab:3DResults} should not be compared.

We perform experiments to clarify the difference between Organ\{A,C,S\}MNIST and OrganMNIST3D. We slice the OrganMNIST3D dataset in the axial / coronal / sagittal views (planes) respectively to generate the central slices. For each view, we take the $60\%$ central slices when slicing and discard the other $40\%$ slices. We evaluate the model performance on the OrganMNIST3D, with 2D-input ResNet-18 trained with Organ\{A,C,S\}MNIST and the axial / coronal / sagittal central slices of OrganMNIST3D, as well as 3D-input ResNet-18. The results are reported in Table~\ref{tab:Organ2/3D_Results}. The performance of 3D-input models is comparable to that of 2D-input models with axial view in general. In other words, with an appropriate setting, the 2D inputs and 3D inputs are comparable on the OrganMNIST3D dataset.

\begin{table*}[!htb]
\footnotesize
	\caption{{\bf Model performance on OrganMNIST3D test set in various settings}, including (upper) 2D-input ResNet-18~\cite{he2016deep} trained with Organ\{A,C,S\}MNIST and axial / coronal / sagittal central slices of OrganMNIST3D, and (lower) 3D-input ResNet-18 with 2.5D / 3D / ACS~\cite{9314699} convolutions, trained with OrganMNIST3D (same as Table~\ref{tab:3DResults}).}
	\label{tab:Organ2/3D_Results}
	\begin{center}
		
		\begin{tabular}{lcc}
			\toprule
			Methods & AUC & ACC \\ 
			\midrule
			\textit{2D-Input ResNet-18}\\
			Trained with OrganAMNIST     & 0.995 & 0.907  \\
			Trained with axial central slices of OrganMNIST3D    & 0.995 & 0.916  \\
			Trained with OrganCMNIST        & 0.991 & 0.877  \\
			Trained with coronal central slices of OrganMNIST3D  & 0.992 & 0.890  \\
			Trained with OrganSMNIST         & 0.959 & 0.697  \\
			Trained with sagittal central slices of OrganMNIST3D & 0.963 & 0.701  \\
			\midrule
			\textit{3D-Input ResNet-18}\\
			2.5D trained with OrganMNIST3D  & 0.977 & 0.788  \\
			3D trained with OrganMNIST3D  & 0.996 & 0.907  \\  
			ACS trained with OrganMNIST3D  & 0.994 & 0.900  \\
			\bottomrule
		\end{tabular}
	\end{center}
\end{table*}

\section*{Usage Notes}

The MedMNIST can be freely available at \url{https://medmnist.com/}. We would be grateful if the users of MedMNIST dataset could cite MedMNIST v1~\cite{medmnistv1} and v2 (this paper), as well as the corresponding source dataset in the publications.

Please note that this dataset is NOT intended for clinical use, as substantially reducing the resolution of medical images might result in images that are insufficient to represent and capture different disease pathologies.

\section*{Code availability}

The data API and evaluation script in Python is available at \url{https://github.com/MedMNIST/MedMNIST}. The reproducible experiment codebase is available at \url{https://github.com/MedMNIST/experiments}.

\section*{Acknowledgements} 

This work was supported by National Science Foundation of China (U20B200011, 61976137). This work was also supported by Grant YG2021ZD18 from Shanghai Jiao Tong University Medical Engineering Cross Research. We would like to acknowledge all authors of the open datasets used in this study.

\section*{Author contributions statement}
JY conceived the experiments.
JY and RS developed the code and benchmark.
JY, RS, DW, ZL, LZ, BK and HP contributed to data collection, cleaning and annotations.
JY, RS, DW and BN wrote the manuscript.
All authors reviewed the manuscript. 

\section*{Competing interests} 
The authors declare no competing interests.

\end{document}